\documentclass[11pt,a4paper]{article}
\usepackage[hyperref]{acl2020}
\usepackage{times}
\usepackage{latexsym}

\usepackage{tgtermes}
\usepackage{amssymb}
 \usepackage{scalefnt,letltxmacro}
\usepackage{mathtools}          
 \LetLtxMacro{\oldtextsc}{\textsc}
 \renewcommand{\textsc}[1]{\oldtextsc{\scalefont{1.10}#1}}
 \usepackage[scaled=0.92]{PTSans}
\usepackage[algoruled]{algorithm2e}
\usepackage{listings}
\usepackage{fancyvrb}
\fvset{fontsize=\normalsize}
\usepackage{setspace}
\SetKwComment{Comment}{$\triangleright$\ }{}
\SetCommentSty{shape}
\usepackage{graphicx}
\usepackage[labelfont=bf, labelsep=period]{caption}
\usepackage[format=hang]{subcaption}
\usepackage{booktabs}
\usepackage[nameinlink]{cleveref}
\usepackage[acronym,smallcaps,nowarn]{glossaries}
\usepackage{framed}

\newcommand{\mba}{\boldsymbol{a}}

\newcommand{\mbw}{\boldsymbol{w}}
\newcommand{\mbx}{\boldsymbol{x}}
\newcommand{\mby}{\boldsymbol{y}}
\newcommand{\mbz}{\boldsymbol{z}}

\newcommand{\mbbeta}{\boldsymbol{\beta}}

\newcommand{\mbeta}{\boldsymbol{\eta}}

\newcommand{\mbmu}{\boldsymbol{\mu}}

\newcommand{\mbsigma}{\boldsymbol{\sigma}}

\newcommand{\mbtheta}{\boldsymbol{\theta}}

\newcommand{\E}{\mathbb{E}}

\newcommand{\cN}{\mathcal{N}}

\newcommand{\N}{\mathcal{N}}

\newcommand{\Gam}{\textrm{Gamma}}

\newcommand{\Pois}{\textrm{Pois}}

 \newacronym{KL}{kl}{Kullback-Leibler}
\newacronym{ELBO}{elbo}{\emph{evidence lower bound}}
\newacronym{POPELBO}{pop-elbo}{\emph{population evidence lower bound}}

\newacronym{SVI}{svi}{stochastic variational inference}
\newacronym{BUMPVI}{bump-vi}{bumping variational inference}

\newacronym{GMM}{gmm}{Gaussian mixture model}
\newacronym{LDA}{lda}{latent Dirichlet allocation}

\newacronym{SUTVA}{sutva}{stable unit treatment value assumption}

\newacronym{TBIP}{tbip}{text-based ideal point model}

\usepackage{microtype}

\aclfinalcopy 
\def\aclpaperid{2410} 

\title{Text-Based Ideal Points}

\author{Keyon Vafa \\
  Columbia University \\
  \textsf{keyon.vafa@columbia.edu} \\\And
  Suresh Naidu \\
  Columbia University \\
  \textsf{sn2430@columbia.edu} \\\And
  David M. Blei \\
  Columbia University \\
  \textsf{david.blei@columbia.edu} \\}

\date{}

\begin{document}
\maketitle
\begin{abstract}
  \noindent
    Ideal point models analyze lawmakers' votes to quantify their
  political positions, or ideal points.
  But votes are not the only way to express a political
  position. Lawmakers also give speeches, release press statements,
  and post tweets. 
  In this paper, we introduce the
  \gls{TBIP}, an unsupervised probabilistic topic model that analyzes texts to
  quantify the political positions of its authors. We demonstrate the
  \gls{TBIP} with two types of politicized text data: U.S. Senate speeches
  and senator tweets.  Though the model does not analyze
  their votes or political affiliations, 
  the \gls{TBIP} separates lawmakers by party, learns
  interpretable politicized topics, and infers ideal points close to
  the classical vote-based ideal points.  One benefit of analyzing
  texts, as opposed to votes, is that the \gls{TBIP} can estimate
  ideal points of anyone who authors political texts, including
  non-voting actors.  To this end, we use it to study tweets from the
  2020 Democratic presidential candidates. Using only the texts of
  their tweets, it identifies them along an interpretable
  progressive-to-moderate \mbox{spectrum}.

 \end{abstract}
\section{Introduction}
\label{sec:intro}

\glsresetall
\begin{figure*}
  \makebox[\textwidth][c]{\includegraphics[width=1.0\textwidth]{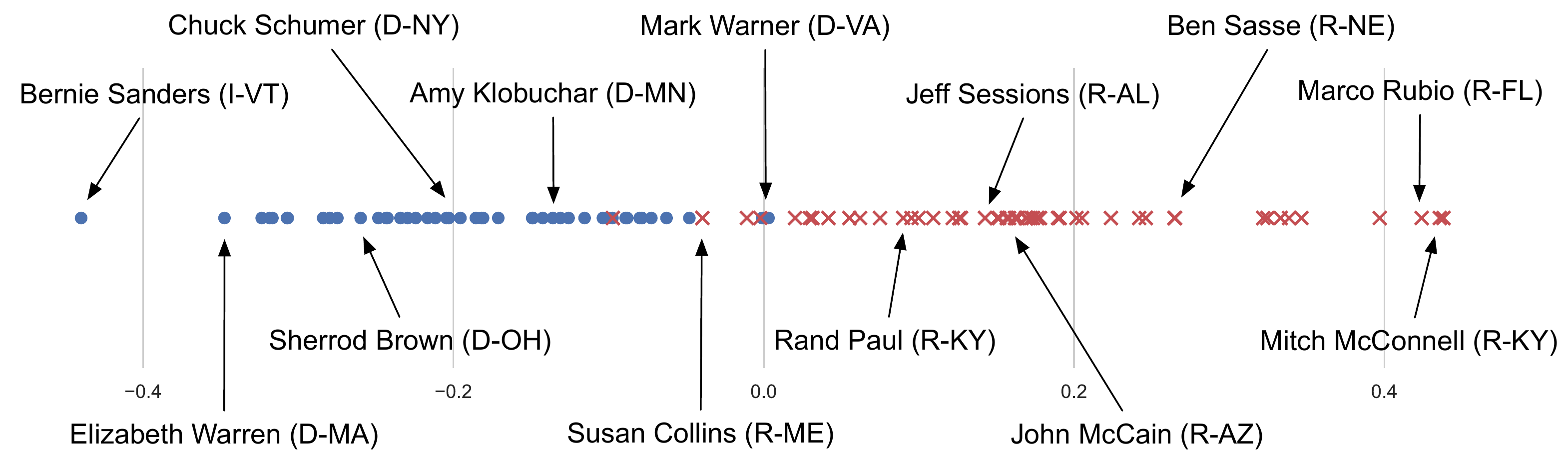}}
  \caption{The \gls{TBIP} separates senators by political party using 
  only speeches. The algorithm does not have access to party information, but senators are coded by their political party for clarity (Democrats in blue circles, Republicans in red x's). The speeches are from the 114th U.S. Senate.}
  \label{fig:senate_speech_ideal_point}
\end{figure*}

\glsresetall
Ideal point models are widely used to help characterize modern
democracies, analyzing lawmakers' votes to estimate their positions on
a political spectrum \citep{poole1985spatial}. But votes aren't the
only way that lawmakers express political preferences---press
releases, tweets, and speeches all help convey their positions.  Like
votes, these signals are recorded and easily collected.

This paper develops the \gls{TBIP}, a probabilistic topic model for
analyzing unstructured political texts to quantify the political
preferences of their authors. While classical ideal point models
analyze how different people vote on a shared set of bills, the
\gls{TBIP} analyzes how different authors write about a shared set of
latent topics. The \gls{TBIP} is inspired by the idea of political framing: 
the specific words and phrases used when discussing a topic can convey
political messages \citep{entman1993framing}. 
Given a corpus of political texts, the \gls{TBIP}
estimates the latent topics under discussion, the latent political
positions of the authors of texts, and how per-topic word choice
changes as a function of the political position of the author. 

A key feature of the \gls{TBIP} is that it is unsupervised. It can be applied
to any political text, regardless of whether the authors belong to known 
political parties. It can also be used to analyze non-voting actors,
such as political candidates.

\Cref{fig:senate_speech_ideal_point} shows a \gls{TBIP} analysis of the speeches of the 114th
U.S. Senate. The model lays the senators out on the real line and
accurately separates them by party. (It does not use party labels in
its analysis.)  Based only on speeches, it has found an interpretable spectrum---Senator Bernie Sanders is liberal, Senator Mitch McConnell is conservative, and Senator Susan Collins is moderate.
For comparison, \Cref{fig:senate_vote_speech_twitter_ideal_point} also shows ideal points estimated
from the voting record of the same senators; their language and their
votes are closely correlated. 

The \gls{TBIP} also finds latent topics, each one a vocabulary-length
vector of intensities, that describe the issues discussed in the
speeches.  For each topic, the \gls{TBIP} involves both a neutral
vector of intensities and a vector of ideological adjustments that
describe how the intensities change as a function of the political
position of the author.  Illustrated in \Cref{tab:ideological_topics_senate_speeches_small} are discovered
topics about immigration, health care, and gun control. In the gun control
topic, the neutral intensities focus on words like ``gun'' and
``firearms.'' As the author's ideal point becomes more negative, terms like ``gun violence'' and ``background checks'' increase in intensity. As the author's ideal point becomes more positive, terms like ``constitutional rights'' increase. 

The \gls{TBIP} is a bag-of-words model that combines ideas from ideal
point models and Poisson factorization topic models
\citep{canny2004gap,gopalan2013scalable}. The latent variables are the
ideal points of the authors, the topics discussed in the corpus, and
how those topics change as a function of ideal point. To approximate the
posterior, we use an efficient black box variational
inference algorithm with stochastic optimization. It scales to large corpora.

We develop the details of the \gls{TBIP} and its variational inference
algorithm.  We study its performance on three sessions of U.S. Senate
speeches, and we compare the \gls{TBIP} to other methods for scaling
political texts~\citep{slapin2008scaling,lauderdale2016measuring}.  The \gls{TBIP} performs
best, recovering ideal points closest to the vote-based ideal points.  We
also study its performance on tweets by U.S. senators, again
finding that it closely recovers their vote-based ideal points.  (In
both speeches and tweets, the differences from vote-based ideal points
are also qualitatively interesting.)  Finally, we study the \gls{TBIP}
on tweets by the 2020 Democratic candidates for President, for which
there are no votes for comparison.  It lays out the candidates along an interpretable progressive-to-moderate spectrum.

 \section{The text-based ideal point model}
\label{sec:methods}

\glsresetall

We develop the \gls{TBIP}, a probabilistic model that infers political
ideology from political texts.  We first review Bayesian ideal points
and Poisson factorization topic models, two probabilistic models on
which the \gls{TBIP} is built.

\subsection{Background: Bayesian ideal points}

Ideal points quantify a lawmaker's political preferences based
on their roll-call votes \citep{poole1985spatial,jackman2001multidimensional,clinton2004statistical}. Consider a group of lawmakers voting ``yea''
or ``nay'' on a shared set of bills. Denote the vote of lawmaker $i$
on bill $j$ by the binary variable $v_{ij}$. The Bayesian ideal point
model posits scalar per-lawmaker latent variables $x_i$ and scalar
per-bill latent variables $(\alpha_j, \eta_j)$.  It assumes the votes
come from a factor model,
\begin{align}
  \label{eqn:ideal_point}
  x_i &\sim \cN(0, 1) \nonumber\\
  \alpha_j, \eta_j &\sim \cN(0, 1) \nonumber\\
  v_{ij} &\sim \text{Bern}(\sigma(\alpha_j + x_i \eta_j)).
\end{align}
where $\sigma(t) = \frac{1}{1+e^{-t}}$.

The latent variable $x_i$ is called the lawmaker's \textit{ideal
  point}; the latent variable $\eta_j$ is the bill's
\textit{polarity}. When $x_i$ and $\eta_j$ have the same sign,
lawmaker $i$ is more likely to vote for bill $j$; when they have
opposite sign, the lawmaker is more likely to vote against it.  The
per-bill intercept term $\alpha_j$ is called the
\textit{popularity}. It captures that some bills are uncontroversial,
where all lawmakers are likely to vote for them (or against them)
regardless of their ideology.

Using data of lawmakers voting on bills, political scientists
approximate the posterior of the Bayesian ideal point model with an
approximate inference method such as Markov Chain Monte Carlo (MCMC)
\citep{jackman2001multidimensional,clinton2004statistical} or
expectation-maximization (EM)~\citep{imai2016fast}. Empirically, the
posterior ideal points of the lawmakers accurately separate political
parties and capture the spectrum of political preferences in American politics \citep{poole2000congress}. 

\subsection{Background: Poisson factorization}

Poisson factorization is a class of non-negative matrix factorization
methods often employed as a topic model for bag-of-words text data
\citep{canny2004gap,cemgil2009bayesian,gopalan2014content}.

Poisson factorization factorizes a matrix of document/word counts into
two positive matrices: a matrix $\mbtheta$ that contains per-document
topic intensities, and a matrix $\mbbeta$ that contains the topics. Denote the count of word $v$ in document
$d$ by $y_{dv}$.  Poisson factorization posits the following
probabilistic model over word counts, where $a$ and $b$ are hyperparameters:
\begin{align}
  \theta_{dk} &\sim \Gam(a, b) \nonumber \\
  \beta_{kv} &\sim \Gam(a, b) \nonumber \\
  \label{eqn:poisson_factorization}
  y_{dv} &\sim \text{Pois}\left(\textstyle \sum_k \theta_{dk}\beta_{kv}\right).
\end{align}
Given a matrix $\mby$, practitioners approximate the posterior
factorization with variational inference \citep{gopalan2013scalable} or MCMC \citep{cemgil2009bayesian}.  Note that Poisson
factorization can be interpreted as a Bayesian variant of nonnegative
matrix factorization, with the so-called ``KL loss function'' \citep{lee1999learning}.

When the shape parameter $a$ is less than 1, the latent vectors
$\theta_d$ and $\beta_k$ tend to be sparse. Consequently, the
marginal likelihood of each count places a high mass around zero and has
heavy tails \citep{ranganath2015deep}.  The posterior
components are interpretable as topics \citep{gopalan2013scalable}.

\begin{table*}
  \small
  \makebox[\textwidth][c]{
  \begin{tabular}{ l  l}
  \multicolumn{1}{c}{Ideology} & \multicolumn{1}{c}{Top Words} \\  \toprule
  \textbf{Liberal} & dreamers, dream, undocumented, daca, comprehensive immigration reform, deport, young, deportation\\
  \textbf{Neutral} & immigration, united states, homeland security, department, executive, presidents, law, country \\
  \textbf{Conservative} & laws, homeland security, law, department, amnesty, referred, enforce, injunction \\ \midrule

  \textbf{Liberal} & affordable care act, seniors, medicare, medicaid, sick, prescription drugs, health insurance, million americans\\
  \textbf{Neutral} &  health care, obamacare, affordable care act, health insurance, insurance, americans, coverage, percent\\
  \textbf{Conservative} & health care law, obamacare, obama, democrats, obamacares, deductibles, broken promises, presidents health care\\ \midrule

  \textbf{Liberal} & gun violence, gun, guns, killed, hands, loophole, background checks, close\\
  \textbf{Neutral} & gun, guns, second, orlando, question, firearms, shooting, background checks\\
  \textbf{Conservative} & second, constitutional rights, rights, due process, gun control, mental health, list, mental illness \\ \bottomrule
 \end{tabular}}
 \caption{The \gls{TBIP} learns topics from Senate speeches that vary as a 
 function of the senator's political positions. 
 The neutral topics are for an ideal point of 0; the ideological topics 
 fix ideal points at $-1$ and $+1$. We interpret one extreme as liberal and
 the other as conservative. Data is from the 114th U.S. Senate.}
 \label{tab:ideological_topics_senate_speeches_small}
 \end{table*}

\subsection{The text-based ideal point model}
\glsresetall

The \gls{TBIP} is a probabilistic model that is designed to infer
political preferences from political texts.

There are important differences between a dataset of votes and a
corpus of authored political language. A vote is one of two choices,
``yea'' or ``nay.'' But political language is high dimensional---a
lawmaker's speech involves a vocabulary of thousands.  A vote sends a
clear signal about a lawmaker's opinion about a bill. But political
speech is noisy---the use of a word might be irrelevant to ideology,
provide only a weak signal about ideology, or change signal 
depending on context. Finally,
votes are organized in a matrix, where each one is unambiguously
attached to a specific bill and nearly all lawmakers vote on all
bills. But political language is unstructured and sparse.  A corpus of
political language can discuss any number of issues---with speeches possibly
involving several issues---and the issues are unlabeled and possibly
unknown in advance.

The \gls{TBIP} is based on the concept of political framing. 
Framing is the idea that a communicator will emphasize 
certain aspects of a message -- implicitly or explicitly -- 
to promote a perspective or agenda \citep{entman1993framing,chong2007framing}.
In politics, an author's word choice for a particular issue
is affected by the ideological message she is trying to convey. A conservative
discussing abortion is more likely to use terms such as ``life'' and 
``unborn,'' while a liberal discussing abortion is more likely to use terms 
like ``choice'' and ``body.'' 
In this example, a conservative is framing
the issue in terms of morality, while a liberal is framing the issue in 
terms of personal liberty.

The \gls{TBIP} casts political framing in a
probabilistic model of language.  While the classical ideal point model infers
ideology from the differences in votes on a shared set of bills, the
\gls{TBIP} infers ideology from the differences in word choice on a
shared set of topics.

The \gls{TBIP} is a probabilistic model that builds on Poisson
factorization.  The observed data are word counts and authors:
$y_{dv}$ is the word count for term $v$ in document $d$, and $a_d$ is
the author of the document.  Some of the latent variables in the
\gls{TBIP} are inherited from Poisson factorization: the non-negative
$K$-vector of per-document topic intensities is $\theta_d$ and the
topics themselves are non-negative $V$-vectors $\beta_k$, where $K$ is the number of topics and $V$ is the vocabulary size. 
We refer to $\mbbeta$ as the \textit{neutral topics}.  Two additional latent variables
capture the politics: the ideal point of an author $s$ is a real-valued
scalar $x_s$, and the \textit{ideological topic} is a real-valued
$V$-vector~$\eta_k$.

The \gls{TBIP} uses its latent variables in a generative model of
authored political text, where the ideological topic adjusts the
neutral topic---and thus the word choice---as a function of the ideal
point of the author.  Place sparse Gamma priors on $\mbtheta$ and~$\mbbeta$, and normal priors on $\mbeta$ and $\mbx$, so for all documents $d$, words $v$,
topics $k$, and authors $s$,
\begin{align*}
  \theta_{dk} &\sim \Gam(a, b) &\eta_{kv} &\sim \cN(0, 1)\\
  \beta_{kv} &\sim \Gam(a, b) & x_s &\sim \cN(0, 1).
\end{align*}
These latent variables interact to draw the count of term $v$ in
document $d$,
\begin{align}
  \label{eqn:tbip}
  y_{dv} & \sim \textstyle
           \Pois\left(\sum_k \theta_{dk} \beta_{kv}
           \exp\{x_{a_d} \eta_{kv}\} \right).
\end{align}
For a topic $k$ and term $v$, a non-zero $\eta_{kv}$ will increase the
Poisson rate of the word count if it shares the same sign as the ideal
point of the author $x_{a_d}$, and decrease the Poisson rate if they
are of opposite signs.  Consider a topic about gun control and suppose
$\eta_{kv}>0$ for the term ``constitution.'' An author with an ideal
point $x_s>0$, say a conservative author, will be more likely to use the
term ``constitution'' when discussing gun control; an author with an
ideal point $x_s < 0$, a liberal author, will be less likely to use
the term.  Suppose $\eta_{kv}<0$ for the term ``violence.'' Now the
liberal author will be more likely than the conservative to use this term.  Finally suppose
$\eta_{kv} = 0$ for the term ``gun.''  This term will be equally
likely to be used by the authors, regardless of their ideal points.

To build more intuition, examine the elements of the sum in the
Poisson rate of \Cref{eqn:tbip} and rewrite slightly to
$\theta_{dk} \exp(\log \beta_{kv} + x_{a_d} \eta_{kv})$. Each of these
elements mimics the classical ideal point model in
\Cref{eqn:ideal_point}, where $\eta_{kv}$ now measures the ``polarity'' of
term $v$ in topic $k$ and $\log \beta_{kv}$ is the intercept or
``popularity.''  When $\eta_{kv}$ and $x_{a_d}$ have the same sign, term $v$ is more likely to be used when discussing topic $k$.  If
$\eta_{kv}$ is near zero, then the term is not politicized, and its
count comes from a Poisson factorization.  For each document $d$, the
elements of the sum that contribute to the overall rate are those for
which $\theta_{dk}$ is positive; that is, those for the topics that
are being discussed in the document.

The posterior distribution of the latent variables provides estimates of the ideal points, neutral topics, and ideological topics. For example, we estimate this posterior distribution using a dataset of senator speeches from the 114th United States Senate session. The fitted ideal points in \Cref{fig:senate_speech_ideal_point} show that the \gls{TBIP} largely separates lawmakers by political party, despite not having access to these labels or votes. \Cref{tab:ideological_topics_senate_speeches_small} depicts neutral topics (fixing the fitted $\hat \eta_{kv}$ to be 0) and the corresponding ideological topics by varying the sign of $\hat \eta_{kv}$. The topic for immigration shows that a liberal framing emphasizes ``Dreamers'' and ``DACA'', while the conservative frame emphasizes ``laws'' and ``homeland security.'' We provide more details and empirical studies in \Cref{sec:empirical_studies}.

 \section{Related work}
\label{sec:related_work}

Most ideal point models focus on legislative roll-call votes. These
are typically latent-space factor models
\citep{poole1985spatial,mccarty1997income,poole2000congress}, which
relate closely to item-response models
\citep{bock1981marginal,bailey2001ideal}. Researchers have also
developed Bayesian analogues
\citep{jackman2001multidimensional,clinton2004statistical} and
extensions to time series, particularly for analyzing the Supreme
Court \citep{martin2002dynamic}.

Some recent models combine text with votes or party information to estimate 
ideal points of
legislators. \citet{gerrish2011predicting} analyze votes and the text
of bills to learn ideological language.  \citet{gerrish2012they},
\citet{lauderdale2014scaling}, and \citet{song2018neural} use text and vote data to learn ideal
points adjusted for topic. The models in
\citet{nguyen2015tea} and \citet{kim2018estimating} analyze votes and floor
speeches together. With labeled political party affiliations, machine learning 
methods can also help 
map language to party membership. \citet{iyyer2014political} use neural
networks to learn partisan phrases, while the models in \citet{tsur2015frame} 
and \citet{gentzkow2019measuring} use political party labels to analyze 
differences in speech patterns. Since the \gls{TBIP} does not use votes
or party information, it is applicable to all political texts, even when votes
and party labels are not present. Moreover, party labels
can be restrictive because they force hard membership in one of two groups (in
American politics). The \gls{TBIP} can infer how topics change smoothly across
the political spectrum, rather than simply learning topics for each political
party.

Annotated text data has also been used to predict ideological positions.
Wordscores 
\citep{laver2003extracting,lowe2008understanding} uses texts that are
hand-labeled by political position to measure the conveyed positions
of unlabeled texts; it has been used to measure the political
landscape of Ireland \citep{benoit2003estimating, herzog2015most}.
\citet{ho2008measuring} analyze hand-labeled editorials to estimate
ideal points for newspapers. The ideological topics learned by the \gls{TBIP} 
are also related to political frames 
\citep{entman1993framing,chong2007framing}. Historically, these frames have 
either been hand-labeled by annotators 
\citep{baumgartner2008decline,card2015media} or used annotated data for 
supervised prediction \citep{johnson-etal-2017-leveraging,baumer2015testing}.
In contrast to these methods, the \gls{TBIP} is completely unsupervised. 
It learns ideological topics that do not need to conform to pre-defined frames.
Moreover, it does not depend on the subjectivity of coders.

\textsc{wordfish} \citep{slapin2008scaling} is a model of authored
political texts about a single issue, similar to a single-topic
version of \gls{TBIP}. \textsc{wordfish} has been applied to party
manifestos \citep{proksch2009avoid,lo2016ideological} and single-issue
dialogue \citep{schwarz2017estimating}.  \textsc{wordshoal} \citep{lauderdale2016measuring} extends
\textsc{wordfish} to multiple issues by analyzing a collection of
labeled texts, such as Senate speeches labeled by debate topic.
\textsc{wordshoal} fits separate \textsc{wordfish} models to the texts
about each label, and combines the fitted models in a one-dimensional
factor analysis to produce ideal points. In contrast to these models, 
the \gls{TBIP} does not require a grouping of the texts into single issues.  
It naturally accommodates unstructured texts, such as tweets, and learns both 
ideal points for the authors and ideology-adjusted topics for the (latent) 
issues under discussion. Furthermore, by relying on stochastic optimization, 
the \gls{TBIP} algorithm scales to large data sets. In
\Cref{sec:empirical_studies} we empirically study how the \gls{TBIP}
ideal points compare to both of these models.

 \section{Inference}
\label{sec:inference}

The \gls{TBIP} involves several types of latent variables: neutral
topics $\beta_k$, ideological topics $\eta_k$, topic intensities
$\theta_d$, and ideal points $x_s$.  Conditional on the text, we
perform inference of the latent variables through the posterior
distribution~$p(\mbtheta, \mbbeta, \mbeta, \mbx | \mby)$. But
calculating this distribution is intractable. We rely on approximate
inference.

We use mean-field variational inference to fit an approximate
posterior distribution
\citep{jordan1999introduction,wainwright2008graphical,blei2017variational}.
Variational inference frames the inference problem as an optimization
problem. Set $q_{\phi}(\mbtheta, \mbbeta, \mbeta, \mbx)$ to be a
variational family of approximate posterior distributions, indexed by
variational parameters $\phi$.  Variational inference aims to find the
setting of $\phi$ that minimizes the KL divergence between $q_\phi$
and the posterior.

Minimizing this KL divergence is equivalent to maximizing the
\gls{ELBO},
\begin{align*}
  \E_{q_{\phi}}[\log p(\mbtheta, \mbbeta, \mbeta, \mbx) &+ \log p(\mby
  |\mbtheta, \mbbeta, \mbeta, \mbx) \nonumber \\
  &- \log q_\phi(\mbtheta, \mbbeta,\mbeta, \mbx)].
\end{align*}
The \gls{ELBO} sums the expectation of the log joint---here broken up
into the log prior and log likelihood---and the entropy of the
variational distribution.

To approximate the \gls{TBIP} posterior we set the variational family
to be the mean-field family.  The mean-field family factorizes over
the latent variables, where $d$ indexes documents, $k$ indexes topics, and 
$s$ indexes authors:
\begin{align*}
  \label{eqn:mean-field}
  q_\phi(\mbtheta, \mbbeta, \mbeta, \mbx) =
  \prod_{d,k,s} q(\theta_d)q(\beta_k)q(\eta_k)q(x_s).
\end{align*}
We use lognormal factors for the positive variables and Gaussian
factors for the real variables,
\begin{align*}
  q(\theta_d) &= \mathrm{LogNormal}_K(\mu_{\theta_d}, I \sigma^2_{\theta_d}) \\
  q(\beta_k) &= \mathrm{LogNormal}_V(\mu_{\beta_k}, I \sigma^2_{\beta_k})\\
  q(\eta_k) &= \N_V(\mu_{\eta_k}, I \sigma^2_{\eta_k}) \\
  q(x_s) &= \N(\mu_{x_s}, \sigma^2_{x_s}).
\end{align*}
Our goal is to optimize the \gls{ELBO} with respect to
$\phi = \{\mbmu_\theta, \mbsigma^2_\theta, \mbmu_\beta,
\mbsigma^2_\beta, \mbmu_\eta, \mbsigma^2_\eta, \mbmu_x,
\mbsigma^2_x\}$.

We use stochastic gradient ascent.  We form noisy gradients with Monte
Carlo and the ``reparameterization trick''
\citep{kingma2013auto,rezende2014stochastic}, as well as with data
subsampling \citep{hoffman2013stochastic}.  To set the step size, we
use Adam \citep{kingma2014adam}.

\begin{figure*}
  \makebox[\textwidth][c]{\includegraphics[width=1.0\textwidth]{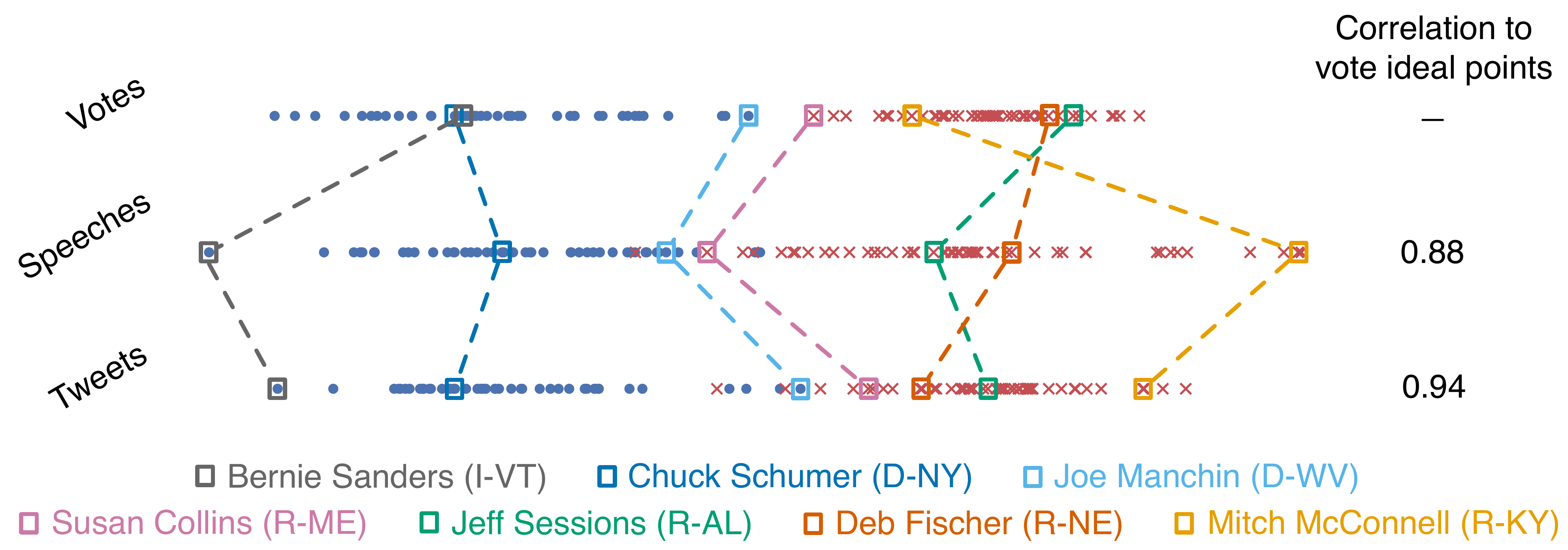}}
  \caption{The ideal points learned by the \gls{TBIP} for senator speeches and 
  tweets are highly correlated with the classical vote ideal points. Senators are coded by their political party (Democrats in blue circles, Republicans in red x's). Although the algorithm does not have access to these labels, the \gls{TBIP} almost completely separates parties.}
  \label{fig:senate_vote_speech_twitter_ideal_point}
\end{figure*}

We initialize the neutral topics and topic intensities with a
pre-trained model.  Specifically, we pre-train a Poisson factorization
topic model using the algorithm in \citet{gopalan2013scalable}. The
\gls{TBIP} algorithm uses the resulting factorization to initialize
the variational parameters for $\theta_d$ and $\beta_k$. The full procedure is described in 
\Cref{sec:appendix_algorithm}.

For the corpus of Senate speeches described in \Cref{sec:methods}, training takes 5 hours on a single NVIDIA Titan V GPU. We have released open source software in Tensorflow and PyTorch.\footnote{\url{http://github.com/keyonvafa/tbip}}

 \section{Empirical studies}
\label{sec:empirical_studies}

\glsresetall
We study the \gls{TBIP} on several datasets of political texts.  We
first use the \gls{TBIP} to analyze speeches and tweets (separately) from
U.S. senators.  For both types of texts, the \gls{TBIP} ideal points,
which are estimated from text, are close to the classical ideal points,
which are estimated from votes.  We also compare the
\gls{TBIP} to existing methods for scaling political
texts~\citep{slapin2008scaling,lauderdale2016measuring}. The \gls{TBIP}
performs
better, finding ideal points closer to the vote-based ideal points.
Finally, we use the \gls{TBIP} to analyze a group that does not vote:
2020 Democratic presidential candidates. Using only tweets, it estimates ideal points for
the candidates on an interpretable progressive-to-moderate spectrum.

\subsection{The \gls{TBIP} on U.S. Senate speeches}

\begin{table*}
  \normalsize
  \setlength{\tabcolsep}{10pt}  
  \renewcommand{\arraystretch}{1.0}  
  \centering
  \begin{tabular}{ l  c c  c c  c c  c c}
  \multicolumn{1}{c}{} & \multicolumn{2}{c}{\textbf{Speeches 111}} &\multicolumn{2}{c}{\textbf{Speeches 112}} & \multicolumn{2}{c}{\textbf{Speeches 113}} & \multicolumn{2}{c}{\textbf{Tweets 114}} \\  \toprule 
  & Corr. & SRC & Corr. & SRC & Corr. & SRC & Corr. & SRC \\
    \midrule
  \textsc{wordfish}    & 0.47 & 0.45  & 0.52 & 0.53 & 0.69 & 0.64& 0.87 & 0.80\\
  \textsc{wordshoal}   & 0.61 & 0.64  & 0.60 & 0.56 & 0.45 & 0.44& --- &  --- \\
  \textsc{tbip}        & \textbf{0.79} & \textbf{0.73}  & \textbf{0.86} & \textbf{0.85} & \textbf{0.87} & \textbf{0.84}& \textbf{0.94} & \textbf{0.84}\\ 
    \bottomrule
 \end{tabular}
 \caption{The \gls{TBIP} learns ideal points most similar to the classical 
 vote ideal points for U.S. senator speeches and tweets. It learns closer ideal points than \textsc{wordfish} and \textsc{wordshoal} in terms of both correlation (Corr.) and Spearman's rank correlation (SRC). The numbers in the column titles refer to the Senate session of the corpus. \textsc{wordshoal} cannot be applied to tweets because there are no debate labels.}
 \label{tab:vote_comparison_scores}
\end{table*}

We analyze Senate speeches provided by \citet{senatespeechesdata},
focusing on the 114th session of Congress (2015-2017). We compare
ideal points found by the \gls{TBIP} to the vote-based ideal point
model from \Cref{eqn:ideal_point}. (\Cref{sec:appendix_data_inference}
provides details about the comparison.) We use approximate posterior
means, learned with variational inference, to estimate the latent variables.  The estimated ideal points are
$\hat{\mbx}$; the estimated neutral topics are $\hat{\mbbeta}$; the
estimated ideological topics are $\hat{\mbeta}$. 

\Cref{fig:senate_vote_speech_twitter_ideal_point} compares the
\gls{TBIP} ideal points on speeches to the vote-based ideal points.\footnote{Throughout our analysis, we appropriately rotate and standardize ideal points so they are visually comparable.} Both models largely separate Democrats and
Republicans. In the \gls{TBIP} estimates, progressive senator Bernie
Sanders (I-VT) is on one extreme, and
Mitch McConnell (R-KY) is on the other. Susan Collins (R-ME), a
Republican senator often described as moderate, is near the middle.
The correlation between the \gls{TBIP} ideal points and vote ideal points
is high, $0.88$. Using only the text of the speeches, the \gls{TBIP}
captures meaningful information about political preferences,
separating the political parties and organizing the lawmakers on a
meaningful political~spectrum.

We next study the topics.  For selected topics,
\Cref{tab:ideological_topics_senate_speeches_small} shows neutral
terms and ideological terms.  To visualize the neutral topics, we list
the top words based on $\hat{\beta}_k$.  To visualize the ideological
topics, we calculate term intensities for two poles of the political
spectrum, $x_s = -1$ and $x_s = +1$. For a fixed $k$, the ideological
topics thus order the words by $\E [\beta_{kv} \exp(-\eta_{kv})]$ and $\E [\beta_{kv} \exp(\eta_{kv})]$.

Based on the separation of political parties in 
\Cref{fig:senate_speech_ideal_point}, we interpret negative ideal points as
liberal and positive ideal points
as conservative. \Cref{tab:ideological_topics_senate_speeches_small} shows that when discussing immigration, a senator with a neutral ideal point uses terms like ``immigration'' and ``United States.'' As the author 
moves left, she will use terms like ``Dreamers'' and ``DACA.'' As she moves 
right, she will emphasize terms like ``laws'' and ``homeland security.''
The \gls{TBIP}
also captures that those on the left refer to health care legislation
as the Affordable Care Act, while those on the right call it
Obamacare. Additionally, a liberal senator discussing guns brings
attention to gun control: ``gun violence'' and ``background checks''
are among the largest intensity terms. Meanwhile, conservative
senators are likely to invoke gun rights, emphasizing ``constitutional rights.''

\paragraph{Comparison to Wordfish and Wordshoal.}  We next treat the
vote-based ideal points as ``ground-truth'' labels and compare the
\gls{TBIP} ideal points to those found by \textsc{wordfish} and
\textsc{wordshoal}. \textsc{wordshoal} requires debate labels, so we
use the labeled Senate speech data provided by \citet{wordshoalreplication}
on the 111th--113th Senates to train each method. 
Because we are interested in comparing models, we use the same 
variational inference procedure to train all methods. 
See \Cref{sec:appendix_data_inference} for more details. 

We use two metrics to compare text-based ideal points to vote-based
ideal points: the correlation between ideal points and Spearman's rank
correlation between their orderings of the senators.  With both
metrics, when compared to vote ideal points from \Cref{eqn:ideal_point}, the
\gls{TBIP} outperforms \textsc{wordfish} and \textsc{wordshoal}; see
\Cref{tab:vote_comparison_scores}.  Comparing to another vote-based 
method, \textsc{dw-nominate} \citep{poole2005spatial}, produces similar results; see \Cref{sec:appendix_dw_nominate}.

\subsection{The \gls{TBIP} on U.S. Senate tweets}

We use the \gls{TBIP} to analyze tweets from U.S. senators during the
114th Senate session, using a corpus provided by \citet{voxgov}. Tweet-based ideal points almost completely separate
Democrats and Republicans; see
\Cref{fig:senate_vote_speech_twitter_ideal_point}. Again, Bernie
Sanders (I-VT) is the most extreme
Democrat, and Mitch McConnell (R-KY) is one of the most extreme Republicans. Susan Collins (R-ME) remains near the middle; she is
among the most moderate senators in vote-based, speech-based, and tweet-based models. The correlation
between vote-based ideal points and tweet-based ideal points is~$0.94$.

We also use senator tweets to compare the \gls{TBIP} to \textsc{wordfish} (we cannot apply \textsc{wordshoal} because tweets do not have debate labels). 
Again, the \gls{TBIP} learns closer ideal points to the classical vote ideal 
points; see \Cref{tab:vote_comparison_scores}.

\subsection{Using the \gls{TBIP} as a descriptive tool}

As a descriptive tool, the \gls{TBIP} provides hints about the
different ways senators use speeches or tweets to convey political
messages.  We use a likelihood ratio to help identify the texts that
influenced the \gls{TBIP} ideal point. Consider the log likelihood of
a document using a fixed ideal point $\tilde x$ and fitted values for
the other latent variables,
\begin{align*}
  \ell_d(\tilde x) =
  \sum_v \log p(y_{dv} | \hat \mbtheta, \hat \mbbeta,
  \hat \mbeta, \tilde x).
\end{align*}
Ratios based on this likelihood can help point to why the \gls{TBIP}
places a lawmaker as extreme or moderate. For a document $d$, if
$\ell_d(\hat x_{a_d}) - \ell_d(0)$ is high then that document was
(statistically) influential in making $\hat x_{a_d}$ more extreme. If
$\ell_d(\hat x_{a_d}) - \ell_d(\max_s(\hat x_s))$ or
$\ell_d(\hat x_{a_d}) - \ell_d(\min_s(\hat x_s))$ is high then that
document was influential in making $\hat x_{a_d}$ less extreme.  We
emphasize this diagnostic does not convey any causal information, but
rather helps understand the relationship between the data and the
\gls{TBIP} inferences.

\begin{figure*}
   \makebox[\textwidth][c]{\includegraphics[width=1.0\textwidth]{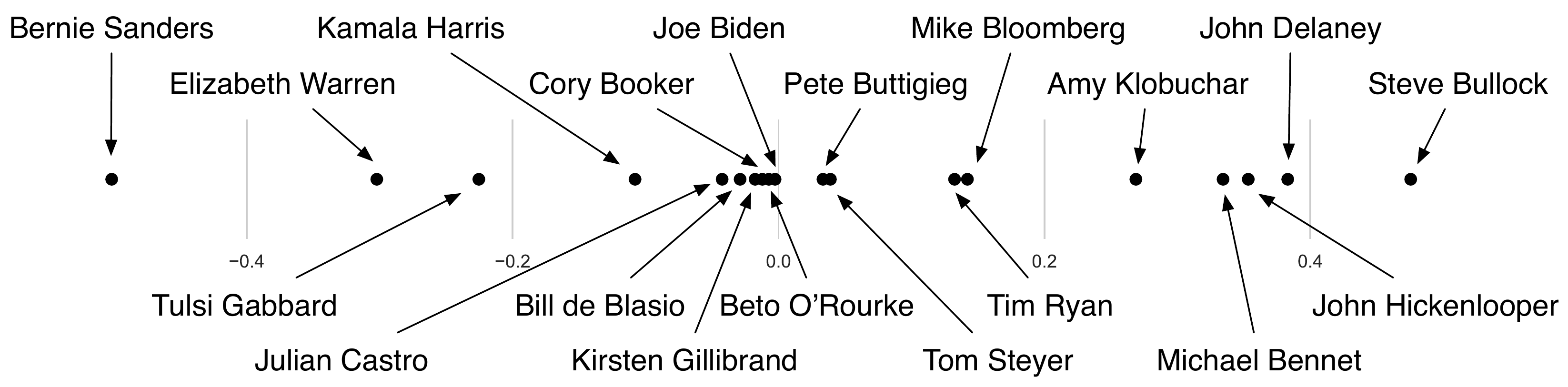}}
  \caption{Based on tweets, the \gls{TBIP} places 2020 Democratic presidential candidates along an interpretable progressive-to-moderate spectrum.}
  \label{fig:candidate_ideal_point}
\end{figure*}

\paragraph{Bernie Sanders (I-VT).} Bernie Sanders is an Independent
senator who caucuses with the Democratic party; we refer to him as a
Democrat. Among Democrats, his ideal point changes the most between
one estimated from speeches and one estimated from votes. Although his
vote-based ideal point is the 17th most liberal, the \gls{TBIP} ideal
point based on Senate speeches is the most extreme.

We use the likelihood ratio to understand this difference in his
vote-based and speech-based ideal points. His speeches with the
highest likelihood ratio are about income inequality and universal
health care, which are both progressive issues. The following is an
excerpt from one such speech:
\begin{quote}
  ``The United States is the only major country on Earth that does not
  guarantee health care to all of our people... At a time when the
  rich are getting richer and the middle class is getting poorer, the
  Republicans take from the middle class and working families to give
  more to the rich and large corporations.''
\end{quote}
Sanders is considered one of the most liberal senators; his extreme
speech ideal point is sensible.

That Sanders' vote-based ideal point is not more extreme appears to be
a limitation of the vote-based method. Applying the likelihood ratio
to votes helps illustrate the issue.  (Here a bill takes the place of
a document.) The ratio identifies H.R. 2048 as influential.  This
bill is a rollback of the Patriot Act that Sanders voted against
because it did not go far enough to reduce federal surveillance
capabilities \citep{sanders2015patriot}. In voting ``nay'', he was joined by one Democrat and 30 Republicans, almost
all of whom voted against the bill because they did not want
surveillance capabilities curtailed at all. Vote-based ideal points,
which only model binary values, cannot capture this nuance in his
opinion. As a result, Sanders' vote-based ideal point is pulled to the
right.

\paragraph{Deb Fischer (R-NE).}  Turning to tweets, Deb Fischer's
tweet-based ideal point is more liberal than her vote-based ideal
point; her vote ideal point is
the 11th most extreme among senators, while her tweet ideal point is the 43rd
most extreme. The likelihood ratio identifies the following tweets as 
responsible for this moderation:

\begin{quote}
  ``I want to empower women to be their own best advocates, secure that they have the tools to negotiate the wages they deserve. \#EqualPay''
\end{quote}
\begin{quote}
  ``FACT: 1963 Equal Pay Act enables women to sue for wage discrimination. \#GetitRight \#EqualPayDay''
\end{quote}
The \gls{TBIP} associates terms about equal pay and women's rights
with liberals. A senator with the most liberal ideal point would
be expected to use the phrase ``\#EqualPay'' 20 times as much as a senator
with the most conservative ideal point and ``women'' 9 times as much, using
the topics in Fischer's first tweet above. Fischer's focus on equal pay
for women moderates her tweet ideal~point.

\begin{table*}
  \small
  \makebox[\textwidth][c]{
  \begin{tabular}{ l  l}
  \multicolumn{1}{c}{Ideology} & \multicolumn{1}{c}{Top Words} \\  \toprule
  \textbf{Progressive} & class, billionaire, billionaires, walmart, wall street, corporate, executives, government\\
  \textbf{Neutral} & economy, pay, trump, business, tax, corporations, americans, billion\\
  \textbf{Moderate} &  trade war, trump, jobs, farmers, economy, economic, tariffs, businesses, promises, job\\ \midrule

  \textbf{Progressive} & \#medicareforall, insurance companies, profit, health care, earth, medical debt, health care system, profits\\
  \textbf{Neutral} & health care, plan, medicare, americans, care, access, housing, millions\\
  \textbf{Moderate} & healthcare, universal healthcare, public option, plan, universal coverage, universal health care, away, choice\\ \midrule

  \textbf{Progressive} & green new deal, fossil fuel industry, fossil fuel, planet, pass, \#greennewdeal, climate crisis, middle ground\\
  \textbf{Neutral} & climate change, climate, climate crisis, plan, planet, crisis, challenges, world\\
  \textbf{Moderate} & solutions, technology, carbon tax, climate change, challenges, climate, negative, durable\\

\bottomrule
 \end{tabular}}
 \caption{
 The \gls{TBIP} learns topics from 2020 Democratic presidential candidate
 tweets that vary as a function of the candidate's political positions. 
 The neutral topics are for an ideal point of 0; the ideological topics 
 fix ideal points at $-1$ and $+1$. We interpret one extreme as progressive and
 the other as moderate.}
 \label{tab:ideological_topics_candidates}
 \end{table*}

\paragraph{Jeff Sessions (R-AL).}
The likelihood ratio can also point to model limitations. Jeff Sessions
is a conservative voter, but the \gls{TBIP} identifies his speeches as
moderate. One of the most influential speeches for his moderate text ideal 
point, as identified by the likelihood ratio, criticizes Deferred Actions for 
Childhood Arrivals (DACA), an immigration policy established by President 
Obama that introduced employment opportunities for undocumented individuals
who arrived as children:
\begin{quote}
``The President of the United States is giving work authorizations to more than 4 million people, and for the most part they are adults. Almost all of them are adults. Even the so-called DACA proportion, many of them are in their thirties. So this is an adult job legalization program.''
\end{quote}
This is a conservative stance against DACA. So why does the \gls{TBIP} identify it as moderate? As depicted in \Cref{tab:ideological_topics_senate_speeches_small}, liberals bring up ``DACA'' when discussing immigration, while conservatives emphasize ``laws'' and ``homeland security.'' The fitted expected count of ``DACA'' using the most liberal ideal point for the topics in the above speech is $1.04$, in contrast to $0.04$ for the most conservative ideal point. Since conservatives do not focus on DACA, Sessions 
even bringing up the program sways his ideal point toward the center. 
Although Sessions refers to DACA disapprovingly, the bag-of-words model cannot capture this negative sentiment.

\subsection{2020 Democratic candidates}

We also analyze tweets from Democratic presidential candidates for the
2020 election. Since all of the candidates running for President do not vote 
on a shared set of issues, their ideal points cannot be estimated using
vote-based methods.

\Cref{fig:candidate_ideal_point} shows tweet-based ideal points for
the 2020 Democratic candidates. 
Elizabeth Warren and Bernie Sanders, who are often considered progressive,
are on one extreme. Steve Bullock and
John Delaney, often considered moderate, are on the other. 
The selected topics in \Cref{tab:ideological_topics_candidates}
showcase this spectrum. 
Candidates with progressive ideal points focus on: billionaires 
and Wall Street when discussing the economy, Medicare for All when
discussing health care, and the Green New Deal when discussing climate change. On the other extreme, candidates with moderate ideal points focus on:
trade wars and farmers when discussing the economy, universal plans for
health care, and technological solutions to climate change.

 \glsresetall
\section{Summary}

We developed the \gls{TBIP}, an ideal point model that analyzes texts
to quantify the political positions of their authors. It estimates the
latent topics of the texts, the ideal points of their authors, and how
each author's political position affects her choice of words
within each topic.  We used the \gls{TBIP} to analyze U.S. Senate
speeches and tweets.  Without analyzing the votes themselves, the
\gls{TBIP} separates lawmakers by party, learns interpretable
politicized topics, and infers ideal points close to the classical
vote-based ideal points.  Moreover, the \gls{TBIP} can estimate ideal
points of anyone who authors political texts, including non-voting
actors.  When used to study tweets from 2020 Democratic
presidential candidates, the \gls{TBIP} identifies them along 
a progressive-to-moderate spectrum.

\paragraph*{Acknowledgments} 
This work is funded by ONR N00014-17-1-2131, ONR N00014-15-1-2209, NIH 1U01MH115727-01, NSF CCF-1740833, DARPA SD2 FA8750-18-C-0130, Amazon,
NVIDIA, and the Simons Foundation. Keyon Vafa is supported by NSF grant DGE-1644869. We thank Voxgov for providing us with senator tweet data. We also thank Mark Arildsen, Naoki Egami, Aaron Schein, and anonymous reviewers for helpful comments and feedback.

\bibliography{writeup}
\bibliographystyle{acl_natbib}

\appendix
\glsresetall

\section{Algorithm}
\label{sec:appendix_algorithm}

\glsreset{TBIP}
\begin{algorithm*}
\SetArgSty{textup}
\DontPrintSemicolon
\setstretch{1.08}
\KwIn{Word counts $\mby$, authors $\mba$, and number of topics $K$ ($D$ documents and $V$ words)}
\KwOut{Document intensities $\hat \mbtheta$, neutral topics $\hat \mbbeta$, ideological topic offsets $\hat \mbeta$, ideal points $\hat \mbx$}
\textbf{Pretrain:} Hierarchical Poisson factorization \citep{gopalan2013scalable} to obtain initial estimates $\hat \mbtheta$ and $\hat\mbbeta$ \\
\textbf{Initialize:} Variational parameters $\mbsigma^2_\theta, \mbsigma^2_\beta,\mbmu_\eta, \mbsigma^2_\eta, \mbmu_x, \mbsigma^2_x$ randomly, $\mbmu_\theta=\log(\hat \mbtheta)$, $\mbmu_\beta = \log(\hat \mbbeta)$\\
Compute weights $\mbw$ as in \Cref{eqn:author_weight}\\
\While{the \gls{ELBO} has not converged} {
  \textbf{sample} a document index $d \in \{1, 2, \dots, D\}$\\
  \textbf{sample} $\mbz_{\theta}, \mbz_\beta, \mbz_\eta, \mbz_x \sim \N(\boldsymbol 0, \boldsymbol I)$
  \Comment*{Sample noise distribution}
  Set $\tilde \mbtheta = \exp(\mbz_\theta \odot \mbsigma_\theta + \mbmu_\theta)$\text{ and }$\tilde \mbbeta = \exp(\mbz_\beta \odot \mbsigma_\beta + \mbmu_\beta)$  
  \Comment*{Reparameterize} 
  Set $\tilde \mbeta = \mbz_\eta \odot \mbsigma_\eta + \mbmu_\eta$\text{ and } $\tilde \mbx = \mbz_x \odot \mbsigma_x + \mbmu_x$ 
  \Comment*{Reparameterize} 
  \For{$v \in \{1, \dots, V\}$}{
    Set $\lambda_{dv} = \left(\sum_k \tilde \theta_{dk}\tilde \beta_{kv}\exp(\tilde  \eta_{kv} \tilde x_{a_d})\right) * w_{a_d}$\\
    Compute $\log p(y_{dv}|\tilde \mbtheta, \tilde \mbbeta, \tilde \mbeta, \tilde \mbx) = \log \text{Pois}(y_{dv}|\lambda_{dv})$ 
    \Comment*{Log-likelihood term}
  }
  Set $\log p(\mby_d | \tilde \mbtheta, \tilde \mbbeta, \tilde \mbeta, \tilde \mbx) = \sum_v \log p(y_{dv} | \tilde \mbtheta, \tilde \mbbeta, \tilde \mbeta, \tilde \mbx)$
  \Comment*{Sum over words}
  Compute $\log p(\tilde \mbtheta, \tilde \mbbeta, \tilde \mbeta, \tilde \mbx)$ and $\log q(\tilde \mbtheta, \tilde \mbbeta, \tilde \mbeta, \tilde \mbx)$
  \Comment*{Prior and entropy terms} 
  Set $\gls{ELBO} = \log p(\tilde \mbtheta, \tilde \mbbeta, \tilde \mbeta, \tilde \mbx) + N * \log p(\mby_d | \tilde \mbtheta, \tilde \mbbeta, \tilde \mbeta, \tilde \mbx)$\\
   $\ \ \ \ \ \ \ \ \ \ \ \ \ \ \ \ \ \ \ \ \ - \log q(\tilde \mbtheta, \tilde \mbbeta, \tilde \mbeta, \tilde \mbx)$\\
  Compute gradients $\nabla_{\phi} \gls{ELBO}$ using automatic differentiation\\
  Update parameters $\phi$\\ 
}
\Return{approximate posterior means $\hat \mbtheta, \hat \mbbeta, \hat \mbeta, \hat \mbx$}
\caption{The \gls{TBIP}}
\label{alg:tbip}
\end{algorithm*}

\glsresetall
We present the full procedure for training the \gls{TBIP} in \Cref{alg:tbip}.
We make a final modification to the model in \Cref{eqn:tbip}. If some political
authors are more verbose than others (i.e. use more words per
document), the learned ideal points may reflect verbosity rather than a political
preference. Thus, we multiply the expected word count by a term that
captures the author's verbosity compared to all authors. Specifically,
if $n_s$ is the average word count over documents for author $s$, we
set a weight:
\begin{equation}
\label{eqn:author_weight}
w_s = \frac{n_s}{\frac{1}{S}\sum_{s'} n_{s'}},
\end{equation}
for $S$ the number of authors.
We then multiply the rate in \Cref{eqn:tbip} by $w_{a_d}$. Empirically, we find this modification does not make much of a difference for the correlation results, but it helps us interpret the ideal points for the qualitative analysis.

\section{Data and inference settings}
\label{sec:appendix_data_inference}
\paragraph{Senator speeches}

We remove senators who made less than 24 speeches. To lessen non-ideological correlations in the speaking patterns of senators from the same state, we remove cities and states in addition to stopwords and procedural terms. We include all unigrams, bigrams, and trigrams that appear in at least 0.1\% of documents and at most 30\%. To ensure that the inferences are not influenced by procedural terms used by a small number of senators with special appointments, we only include phrases that are spoken by 10 or more senators. This preprocessing leaves us with 19,009 documents from 99 senators, along with 14,503 terms in the vocabulary.

To train the \gls{TBIP}, we perform stochastic gradient ascent using Adam \citep{kingma2014adam}, with a mini-batch size of 512. To curtail extreme word count values from long speeches, we take the natural logarithm of the counts matrix before performing inference (appropriately adding 1 and rounding so that a word count of 1 is transformed to still be 1). We use a single Monte Carlo sample to approximate the gradient of each batch. We assume 50 latent topics and posit the following prior distributions: $\theta_{dk}, \beta_{kv} \sim \text{Gamma}(0.3, 0.3)$, $\eta_{kv}, x_s \sim \mathcal{N}(0, 1)$.

We train the vote ideal point model by removing all votes that are not cast as ``yea'' or ``nay'' and performing mean-field variational inference with Gaussian variational distributions. Since each variational family is Gaussian, we approximate gradients using the reparameterization trick \citep{rezende2014stochastic,kingma2014adam}. 

\begin{table*}
  \normalsize
  \setlength{\tabcolsep}{10pt}  
  \renewcommand{\arraystretch}{1.0}  
  \centering
  \begin{tabular}{ l  c c  c c  c c  c c}
  \multicolumn{1}{c}{} & \multicolumn{2}{c}{\textbf{Speeches 111}} &\multicolumn{2}{c}{\textbf{Speeches 112}} & \multicolumn{2}{c}{\textbf{Speeches 113}} & \multicolumn{2}{c}{\textbf{Tweets 114}} \\  \toprule 
  & Corr. & SRC & Corr. & SRC & Corr. & SRC & Corr. & SRC \\
    \midrule
  \textsc{wordfish}    & 0.52 & 0.49  & 0.51 & 0.51 & 0.71 & 0.65 & 0.79 & 0.74\\
  \textsc{wordshoal}   & 0.62 & 0.66  & 0.58 & 0.51 & 0.46 & 0.46 & --- & --- \\
  \textsc{tbip}        & \textbf{0.82} & \textbf{0.77}  & \textbf{0.85} & \textbf{0.85} & \textbf{0.89} & \textbf{0.86}& \textbf{0.94} & \textbf{0.88}\\ 
    \bottomrule
 \end{tabular}
 \caption{
 The \gls{TBIP} learns ideal points most similar to \textsc{dw-nominate} vote ideal points for U.S. senator speeches and tweets. It learns closer ideal points than \textsc{wordfish} and \textsc{wordshoal} in terms of both correlation (Corr.) and Spearman's rank correlation (SRC). The numbers in the column titles refer to the Senate session of the corpus. \textsc{wordshoal} cannot be applied to tweets because there are no debate labels.}
 \label{tab:dw_vote_comparison_scores}
\end{table*}

For the comparisons against \textsc{wordfish} and \textsc{wordshoal}, we preprocess speeches in the same way as \citet{lauderdale2016measuring}. We train each Senate session separately, thereby only including one timestep for \textsc{wordfish}. For this reason, our results on the U.S. Senate differ from those reported by \citet{lauderdale2016measuring}, who train a model jointly over all time periods. Additionally, we use variational inference with reparameterization gradients to train all methods. Specifically, we perform
mean-field variational inference, positing Gaussian variational families on
all real variables and lognormal variational families on all positive 
variables.

\paragraph{Senator tweets}
Our Senate tweet preprocessing is similar to the Senate speech preprocessing, although we now include all terms that appear in at least 0.05\% of documents rather than 0.01\% to account for the shorter tweet lengths. We remove cities and states in addition to stopwords and the names of politicians. This preprocessing leaves us with 209,779 tweets. We use the same model and hyperparameters as for speeches, although we no longer take the natural logarithm of the counts matrix since individual tweets cannot have extreme word counts due to the character limit. We use a batch size of~1,024.

\paragraph{2020 Democratic candidates}
We scrape the Twitter feeds of 19 candidates, including all tweets between January 1, 2019 and February 27, 2020. We do not include Andrew Yang, Jay Inslee, and Marianne Williamson since it is difficult to define the political preferences of non-traditional or single-issue candidates. We follow the same preprocessing we used for the 114th Senate, except we include tokens that are used in more than 0.05\% of documents rather than 0.1\%. We remove phrases used by only one candidate, along with stopwords and candidate names. This preprocessing leaves us with 45,927 tweets for the 19 candidates. We use the same model and hyperparameters as for senator~tweets.

\section{Comparison to DW-Nominate}
\label{sec:appendix_dw_nominate}
\textsc{dw-nominate} \citep{poole2005spatial} is a dynamic method for learning
ideal points from votes. As opposed to the vote ideal point model in 
\Cref{eqn:ideal_point}, it
analyzes votes across multiple Senate sessions. It also
learns two latent dimensions per legislator. We also compare text ideal points to 
the first dimension of DW-Nominate, which corresponds to 
economic/redistributive preferences \citep{boche2018new}. We use the fitted 
\textsc{dw-nominate} ideal points available on Voteview \citep{boche2018new}. 
The \gls{TBIP} learns ideal points closer to \textsc{dw-nominate} than 
\textsc{wordfish} and \textsc{wordshoal}; see \Cref{tab:dw_vote_comparison_scores}.

In \Cref{sec:empirical_studies}, we observed that Bernie Sanders' vote
ideal point is somewhat moderate under the scalar ideal point model
from \Cref{eqn:ideal_point}. It is worth noting that Sanders' vote ideal point
is more extreme
under \textsc{dw-nominate} than under the scalar model: his \textsc{dw-nominate} ideal
point is the third-most extreme among Democrats. Since \textsc{dw-nominate} uses two
dimensions to model each legislator's latent preferences, it can
more flexibly model Sanders' voting deviations. Additionally,
the dynamic nature of \textsc{dw-nominate} may capture salient information from 
other Senate sessions. However, restricting the vote ideal point to be
static and a scalar, like it is for the \gls{TBIP}, results in the
more moderate vote ideal point in \Cref{sec:empirical_studies}.

\end{document}